\pdfoutput=1

\documentclass[11pt]{article}

\usepackage[]{EMNLP2022}

\usepackage{times}
\usepackage{latexsym}

\usepackage[T1]{fontenc}

\usepackage[utf8]{inputenc}

\usepackage{microtype}

\usepackage{inconsolata}

\usepackage{microtype}
\usepackage{balance}
\usepackage[flushleft]{threeparttable}
\usepackage{enumitem}
\usepackage[linesnumbered,ruled,vlined]{algorithm2e}
\usepackage{algpseudocode}
\usepackage{booktabs}
\usepackage{subcaption}
\usepackage[nointegrals]{wasysym}
\usepackage{tikz}
\usepackage{tabu}
\usepackage{tabularx}
\usepackage{float}
\usepackage{listings}
\usepackage{color}
\usepackage{multirow}
\usepackage{framed}
\usepackage{pifont}
\usepackage{dsfont}
\usepackage{xcolor}
\usepackage{amsmath,amsthm}
\usepackage{amssymb}
\usepackage{graphicx}
\usepackage{textcomp}
\usepackage{tcolorbox}
\usepackage{soul}
\usepackage{makecell}
\usepackage[normalem]{ulem}
\usepackage{xspace}
\usepackage{arydshln}

\newcommand{\keywordCode}[1]{{\small \texttt{#1}}}

\newcommand{\modelname}{\mbox{\textsc{DecompT5}}}
\newcommand{\pipelinename}{\mbox{\textsc{DecompEntail}}}

\definecolor{explicit-color}{rgb}{0.88, 0.66, 0.37}
\DeclareRobustCommand{\hlerror}[1]{{\sethlcolor{explicit-color}\hl{#1}}}

\title{ \vspace*{-0.5in}

{{\small \hfill EMNLP'22}\\

\vspace*{.25in}} Learning to Decompose: Hypothetical Question Decomposition Based on Comparable Texts}

\author{Ben Zhou$^{1}$\thanks{~~Work partly done when interning at AI2.} ~ Kyle Richardson$^2$ ~ Xiaodong Yu$^1$ ~ Dan Roth$^1$ \\
{ $^1$University of Pennsylvania \;\;\;  $^2$Allen Institute for AI} \\
{\{xyzhou, xdyu, danroth\}@seas.upenn.edu \;\; kyler@allenai.org}
}

\begin{document}
\maketitle
\begin{abstract}

Explicit decomposition modeling, which involves breaking down complex tasks into more straightforward and often more interpretable sub-tasks, has long been a central theme in developing robust and interpretable NLU systems. However, despite the many datasets and resources built as part of this effort, the majority have small-scale annotations and limited scope, which is insufficient to solve general decomposition tasks.
In this paper, we look at large-scale intermediate pre-training of decomposition-based transformers using distant supervision from comparable texts, particularly large-scale parallel news. We show that with such intermediate pre-training, developing robust decomposition-based models for a diverse range of tasks becomes more feasible. For example, on semantic parsing, our model, \modelname{}, improves 20\% to 30\% on two datasets, Overnight and TORQUE, over the baseline language model. We further use \modelname{} to build a novel decomposition-based QA system named \pipelinename{}, improving over state-of-the-art models, including GPT-3, on both HotpotQA and StrategyQA by 8\% and 4\%, respectively.

\end{abstract}

\section{Introduction}

\begin{figure}[t]
    \centering
    \includegraphics[scale=0.37]{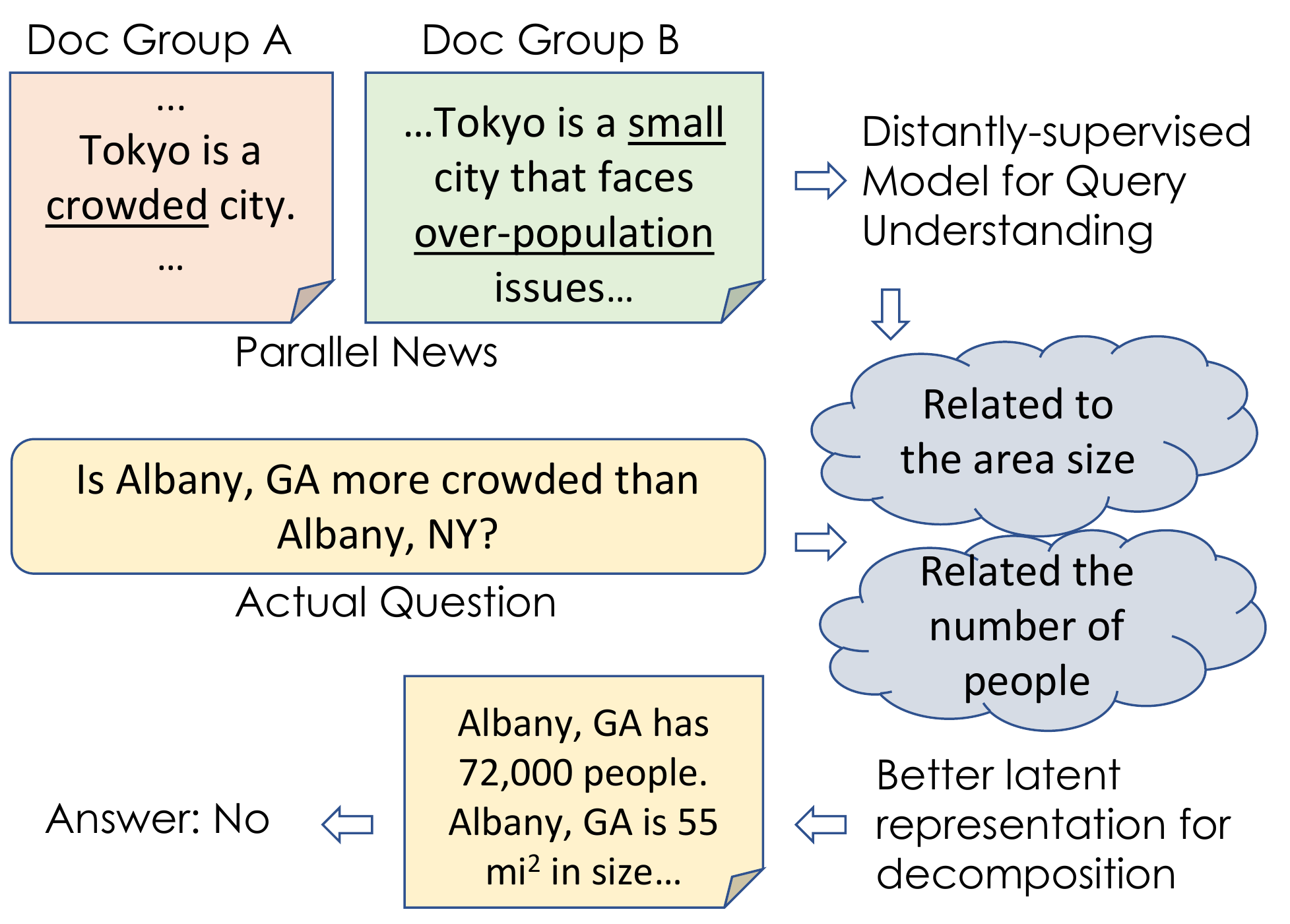}
    \caption{
        An example of how parallel news documents can be used to train a model that is capable of making educated guesses on what the question is asking, and how it may help to derive a better answer.
    }
    \label{fig:overview}
\end{figure}

Answering questions often involves making educated guesses: we do not necessarily have accurate facts but can use common sense to understand what most questions are asking and what kinds of knowledge are needed. For example (see Fig.~\ref{fig:overview}), we can understand the question ``\emph{Is Albany, GA more crowded than Albany, NY?}'' involves comparing the size and population of two cities without knowing the specific numbers to compare.
It is often desirable to make such a decomposition because a city's population is usually much easier to acquire than a direct answer to the original question. 

Existing approaches to end-to-end question-answering (QA) assume that pre-trained language models (LMs) are capable of both robust question understanding of this type and acquiring the relevant facts. Much recent evidence, however, has revealed limitations in the commonsense and compositional reasoning abilities of current transformers \cite{zhou2019going,liu2021challenges}, in part due to \emph{reporting biases} (e.g., relating the semantics of ``more crowded'' and ``overpopulation'' can be difficult given that such contexts rarely co-occur in single document on which models are pre-trained) and other \emph{dataset artifacts} \cite{gururangan2018annotation}. This is even more evident in recent datasets with complex questions that are designed to require decomposition. For example, GPT-3 \cite{Brown2020LanguageMA}, a language model with 175 billion parameters, only achieves mid-60s accuracy on StrategyQA \cite{geva2021did}, a binary QA benchmark with a random baseline at around 50. Moreover, such datasets are often small in size and scope, which makes it difficult to overcome knowledge gaps in LMs through fine-tuning and developing general-purpose decomposition models. 



In this paper,\footnote{\url{http://cogcomp.org/page/publication_view/992}} we attempt to bridge the gap of reporting biases, which hinders LMs from learning implicit connections between questions and decompositions (e.g., ``crowded'' and ``population''). We do this through intermediate pre-training on distant supervision, following recent attempts to distill common sense into transformers via distant supervision \cite{zhou2021temporal}.
Specifically, we use collections of article pairs with parallel descriptions of similar news events from different angles as our distant supervision. As illustrated in Fig.~\ref{fig:overview}, large collections of comparable texts (\S\ref{sec:distant-supervision-intuitions}) contain a wide variety of commonsense implications needed for decomposition. We extract 2.6 million sentence pairs (\S\ref{sec:pl-extraction}) for this purpose, and then train \modelname{} (\S\ref{sec:pretrain}), a T5 \cite{Raffel2020ExploringTL} model that is further-pre-trained on our distant supervision instances. In \S\ref{sec:intrinsic-experiments}, we show that \modelname{}, while simple, serves as a more effective model than the base language model on general question understanding through experiments on Overnight \cite{Wang2015BuildingAS} and TORQUE \cite{Ning2020TORQUEAR} semantic parsing tasks, achieving 22-32\% absolute improvements. 

Since smaller language models cannot sufficiently memorize facts (e.g., the exact population of Albany), they are often used in conjunction with external knowledge retrieval for more complicated tasks such as QA. To bridge this gap, we design a novel QA pipeline using \modelname{} at its core (\S\ref{sec:e2e-qa-system}). The full model and pipeline, called \pipelinename{}, first generates explicit question decompositions, then makes factual corrections on the decomposed statements with GPT-3. As a final step, \pipelinename{} employs an entailment model that derives the final answer with the generated decomposition as the premise and the question and candidate answer as the hypothesis. 

In \S\ref{sec:e2e-experiments}, we show that \pipelinename{}, despite its relatively small size, can generate good decomposition chains and outperforms GPT-3 on both StrategyQA and a binary portion of HotpotQA by 4\% and 8\%, respectively.
This shows that we can improve baseline language models or even much larger reasoners with explicit decomposition, which has the advantage of enhanced interpretability and transferability. On the other hand, \modelname{} only relies on supporting fact annotations instead of explicit reasoning steps, which is more common in datasets and can be better applied for joint learning.

\noindent \textbf{Contributions.} In summary, our contributions are three-fold: 1) we collect distant supervision from parallel news to encourage robust semantic understanding for question decomposition, 2) we train a general decomposition model called \modelname{} with our collected distant supervision that significantly improves over the baseline language models on intrinsic evaluations, and 3) we propose a decomposition-based QA pipeline called \pipelinename{} that relies on \modelname{} at its core. We show that \pipelinename{} has improved performance over several baselines on decomposition-based QA.

\section{Related Work}

Our work relates to the literature on multi-hop reasoning \cite{yang2018hotpotqa}, which has recently produced new annotation schemes (e.g., \emph{QDMR} from \citet{wolfson2020break} and \emph{strategy question decomposition} annotations from \citet{geva2021did}) and datasets for complex reasoning that target explicit model decomposition \cite{talmor2018web,wolfson2020break,geva2021did,khot2022hey}. We take inspiration from systems that build explicit reasoning paths, such as semantic parsers \cite{Liang2011LearningDC, Berant2013SemanticPO}, and their modern variations \cite{Andreas2016NeuralMN,Gupta2020NeuralMN,khot2020text}. 
\citet{Min2019MultihopRC,Perez2020UnsupervisedQD} aim to build general question decomposition models, however, focusing on simpler tasks than our study.  


Our work is also related to sentence-pair datasets collected from comparable texts \cite{Fader2013ParaphraseDrivenLF, Zhang2019PAWSPA, Reimers2019SentenceBERTSE}. Compared to most of these works, our extraction does not use human annotation, and produces clean and diverse signals for question understanding.

Previous work has also discussed using large-scale further pre-training to improve language models \cite{Zhou2020TemporalCS, zhou2021temporal, zhao2021effective}. We follow a similar general scheme with novel extraction sources and focus on a general representation for questions, which resembles some idea in existing work \cite{Khashabi2020UnifiedQACF}.


\section{Distant Supervision for Decomposition}
In \S\ref{sec:distant-supervision-intuitions}, we describe our intuitions on why question decomposition is important and what is missing from existing pre-trained language models for them to do well. Following that, we describe how we collect distant supervision signals to improve the process of learning to decompose in \S\ref{sec:pl-extraction}. In \S\ref{sec:pretrain}, we propose \modelname{}, a T5-based model that is further pre-trained on the collected distant supervision using standard seq-to-seq training objectives.

\subsection{Intuitions}
\label{sec:distant-supervision-intuitions}
\noindent \textbf{Educated Guesses in QA.} We, as humans, need to answer questions all the time, but we may not possess all the facts. For example, an ordinary person may not know the exact populations of Albany to answer ``Is Albany, GA more crowded than Albany, NY'', or the density of corgis to answer `Will a corgi float on water''. However, that person may search for ``population'' or ``density'' instead of the original question to find the answer because we know that it is much easier to find the ``population of a city'' than to find an answer to the original question. The human capacity for guessing what the question is asking and how that question can be decomposed to simpler concepts by associating \textit{crowded} with \textit{population}, and \textit{float} with \textit{density} is crucial for solving day-to-day tasks. However, making such connections can be very challenging for pre-trained language models because of reporting biases. Written texts rarely make such connections explicit in single documents, as most authors expect readers to make many trivial inferences.

\noindent \textbf{Parallel News.} In this work, we aim to bridge this decomposition gap in pre-trained language models through incidental supervision \cite{Roth2017IncidentalSM} from comparable corpora \cite{Klementiev2006WeaklySN}. We find news articles reporting the same news event but from different authors and angles. Related sentences in such parallel news often complement each other and provide new information. This complementary information is often more sophisticated and diverse than paraphrasing, because it contains implications and causal relations. Fig.~\ref{fig:overview} shows an example of how a pair of articles describing Tokyo from slightly different angles may help decompose the running example question. One article mentions that Tokyo is crowded, while the other expresses similar points but focuses on area size and population descriptions. Intuitively, a model may benefit from such connections to learn that ``crowded'' is related to ``size'' and ``count''. It is rare, however, for a single document to contain both aspects, causing difficulties for LMs that primarily learn from single documents.

\subsection{Parallel News Extraction}
\label{sec:pl-extraction}
We use the RealNews corpus \cite{Zellers2019DefendingAN} as the source corpus because it contains cleaned, date-marked new articles from diverse domains. We aim to select news article pairs that describe the same main event and find sentence pairs within these document pairs that are likely to contain complementary information to each other.

\noindent \textbf{Filter Article Pairs.} We select article pairs within a 2-day window of publication because the same news events are typically covered within a relatively short period. We then employ a pre-trained entailment model from SentenceBert \cite{Reimers2019SentenceBERTSE} to check the titles of each article pair and retain those pairs whose titles have a cosine similarity greater than $0.8$.

\noindent \textbf{Find Sentence Pairs.} We then find sentence pairs across each selected article pairs that are related and complementary to each other. To do this, we run the same sentence similarity model and retain all sentence pairs with a similarity score between $0.6$ and $0.9$. The lower bound is to make sure the sentences are approximately related. Even though $0.6$ is considerably a loose bound for many tasks (e.g., paraphrasing), it is suitable in our case because we have a strong assumption that the articles are closely related because of date and title similarities. This lower bound is sufficient to guarantee that the vast majority of sentence pairs above this threshold contain complementary information to each other. For example, the similarity score between ``The US Military has already started withdrawal from Syria'' and ``The US is only moving non-essential equipment out of Syria, because precipitous withdrawal would shatter US policy in Syria and allow IS to rebuild'' is only $0.6$. However, the second sentence provides non-paraphrasing but complementary information to the first sentence. A model may learn that troops in other countries are linked with foreign policy, which is the type of information that is often implicit in single documents. The upper-bound $0.9$ is to filter out sentence pairs that are too similar or simply paraphrasing each other, as these pairs do not provide much additional information to facilitate question understanding. 

\noindent \textbf{Filtering with tf-idf.} We employ an additional filtering process based on sentence topics to keep the final dataset's diversity. To do this, we calculate the inverse document frequency (idf) of each word in the vocabulary based on Wikipedia and multiply that with the term frequency (tf) of each word within the sentence pairs. Next, we use the top three words ranked by td-idf scores of each sentence pair as the ``signature'' and randomly keep ten sentence pairs with identical signatures at most. 2.6 million sentence pairs remain after this step. Finally, we format the data as a standard seq-to-seq training task, where the input sentence is one of the sentences in the pair, while the model is trained to generate the other sentence in the pair. The order is randomly decided. 

\noindent \textbf{Data for Language Modeling Objective.} Beyond the sentence pairs, we also inject some data from Project Gutenberg\footnote{\url{https://www.gutenberg.org/}} and format it to the language model pre-training format (e.g., the denoising objective for T5 \cite{Raffel2020ExploringTL}). We sample around 900K sentences for this purpose.

\subsection{Comparisons with Similar Data Sources}
\begin{table}[t]
\centering
\small
\begin{tabular}{lccccc}
\toprule
Metric / Data & Ours & P-auto & P-inc. & NLI & QA \\
\cmidrule(lr){1-1}\cmidrule(lr){2-2}\cmidrule(lr){3-3}\cmidrule(lr){4-4}\cmidrule(lr){5-5}\cmidrule(lr){6-6}
Length $\uparrow$ & 52 & 42 & 20 & 31 & 40\\ 
Length-diff $\uparrow$ & 9 & 1 & 2 & 10 & 20\\
Embed-sim $\downarrow$ & 0.7 & 1.0 & 0.9 & 0.6 & 0.6\\
Sem-sim $\downarrow$ & 0.7 & 1.0 & 0.9 & 0.7 & 0.7\\
Cost $\downarrow$ & low & low & mid & high & high \\
\bottomrule
\end{tabular}
\caption{Comparisons between our data and other sources for reasoning tasks. \textit{P-auto} is paraphrasing data from automatic (distant) collection, \textit{P-inc.} is paraphrasing data from incidental supervision. \textit{Sem-sim} is semantic similarity. $\uparrow$/$\downarrow$ marks the direction for each metric to present a more diverse data source.}
\label{tab:naive-comparison}
\end{table}
We compare our data collected in \S\ref{sec:pl-extraction} with other sources that may similarly be used, including paraphrasing, textual entailment (NLI), and question-answering (QA). Paraphrasing data can be collected either automatically (e.g., PAWS \cite{Zhang2019PAWSPA}), or from incidental but human-involved processes (e.g., Quora duplicated questions\footnote{\url{https://quoradata.quora.com/}}). We use these two datasets to represent each category respectively. In addition, we use the MNLI dataset \cite{Williams2018ABC} for NLI, and StrategyQA (question+answer/supporting-facts) for QA. We randomly sample 10k sentence pairs from each source. We compare basic statistics, including sentence pair length and the length difference between the two sentences. We also compare sentence similarity via averaged word embeddings \cite{Pennington2014GloVeGV} and sentence-level semantic embeddings \cite{Reimers2019SentenceBERTSE}.\footnote{We use the ``average\_word\_embeddings\_glove.840B.300d'' and ``all-MiniLM-L6-v2'' models, respectively.}
Table~\ref{tab:naive-comparison} shows that our data source provides richer and more diverse information while not requiring any human annotation. This observation aligns with our intuitions in \S\ref{sec:distant-supervision-intuitions}.

\subsection{Pre-training with Distant Supervision}
\label{sec:pretrain}

We use T5-large \cite{Raffel2020ExploringTL} as our base language model due to its sequence-to-sequence architecture and relatively small parameter size (containing 770m parameters) for easier pre-training. We train the base language model on our distant supervision dataset for one epoch and call the resulting model \modelname{}. We expect, however, that this pre-training technique with our collected dataset is beneficial to most existing pre-trained language models, as it bridges the reporting bias gap in general language modeling objectives.

\section{Decomposition-based QA Pipeline}
\label{sec:e2e-qa-system}


Our proposed model \modelname{} has two uses: it can be \textbf{directly fine-tuned} on tasks that require query understanding and decomposition, as we later show in \S\ref{sec:intrinsic-experiments}. It can also be applied in a pipeline fashion to \textbf{produce meaningful decompositions} that help with more complicated tasks that require external knowledge, such as general question answering. This section focuses on the design challenges and choices for such a QA pipeline. We first explain the intuitions in \S\ref{sec:pipeline-intuition}, then describe and propose \pipelinename{} in \S\ref{sec:pipeline}. We evaluate our proposed QA pipeline in \S\ref{sec:e2e-experiments}.

\subsection{Intuitions and Design Choices}
\label{sec:pipeline-intuition}

\begin{figure}[t]
    \centering
    \includegraphics[scale=0.37]{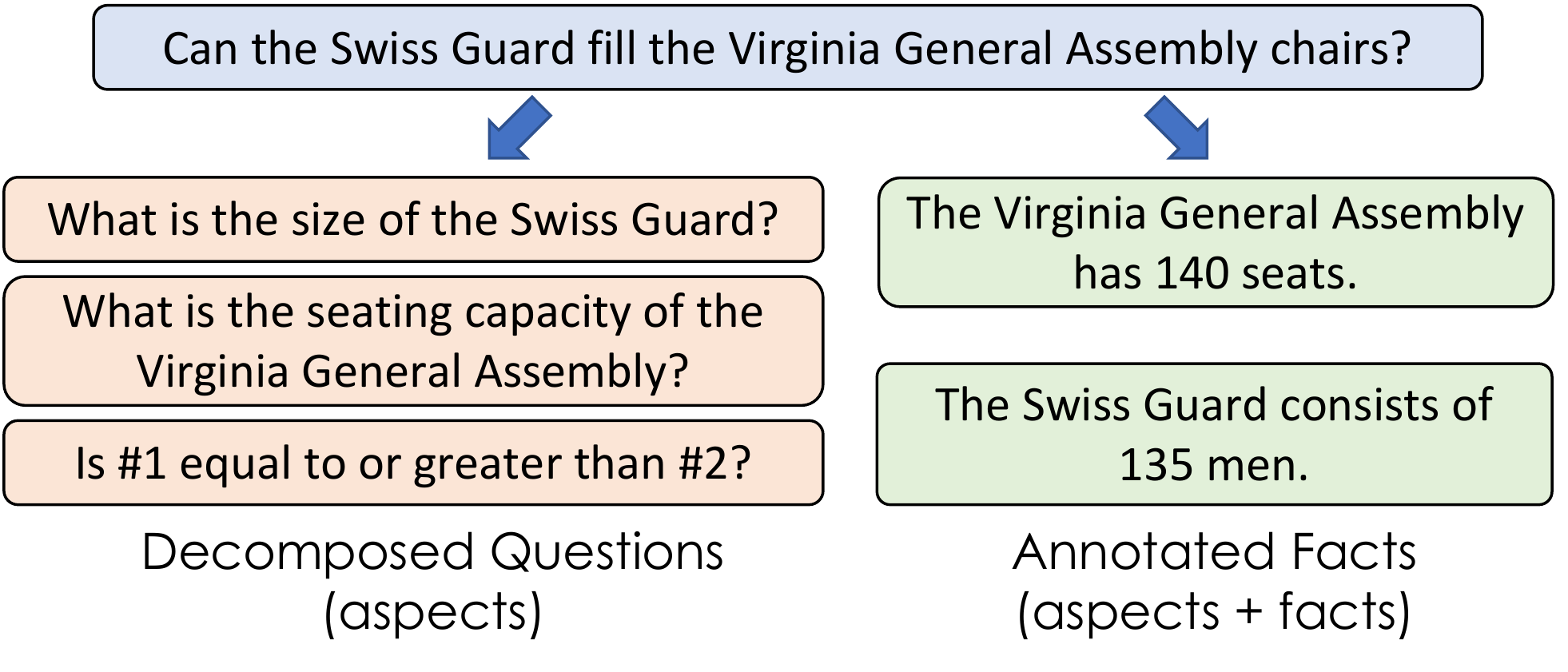}
    \caption{
        An example StrategyQA \citep{geva2021did} instance that includes a question annotated with decomposed questions and their corresponding facts.
    }
    \label{fig:stqa-example}
\end{figure}

\begin{table}[t]
\centering
\small
\begin{tabular}{lcc}
\toprule
Additional Information & \#Train & Accuracy \\
\cmidrule(lr){1-1}\cmidrule(lr){2-2}\cmidrule(lr){3-3}
None & 2061 & 53.3 \\
Aspects & 2061 & 59.7 \\
Facts (Impossible) & n/a & n/a \\
Aspects, Facts & 2061 & 84.5 \\
Aspects, Facts, Indirect & 15296 & \textbf{90.2} \\
\bottomrule
\end{tabular}
\caption{T5-3B accuracy on StrategyQA dev set when different information is provided for both training and evaluation. \textit{Aspects} refers to the knowledge dimensions (without values) that are involved with each question. \textit{Facts} refers to the actual knowledge involved, which is not possible to acquire without knowing the corresponding aspects. \textit{Indirect} contains additional supervision of paraphrasing and entailment. Details are in \S\ref{sec:sanity_check}.}
\label{tab:sanity_check}
\end{table}

\begin{figure*}[t]
    \centering
    \includegraphics[scale=0.5]{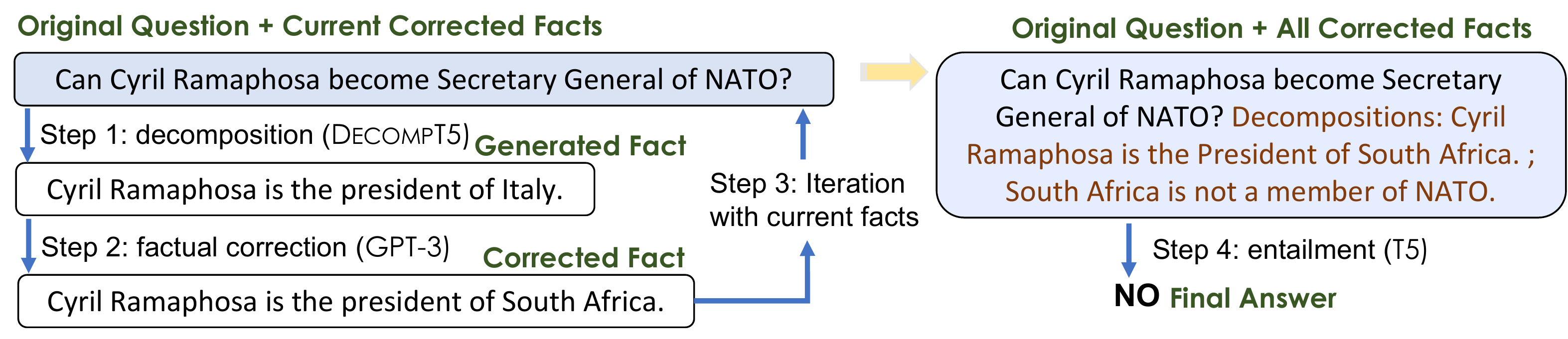}
    \caption{
        An overview of our proposed \pipelinename{} pipeline. The final decomposition is an actual output from the pipeline. See more examples in Fig.~\ref{fig:manual-analysis-examples}. 
    }
    \label{fig:pipeline-overview}
\end{figure*}

As we argue in \S\ref{sec:distant-supervision-intuitions}, an agent can decompose complex questions into simpler and more controlled forms by linking a question to all relevant \textit{aspects} of that question (e.g., the relevant sub-queries related to the input question). With such aspects or components, the agent can make easier knowledge retrieval to acquire the specific values of the aspects, which we call relevant \textit{facts}. As shown in Fig.~\ref{fig:stqa-example}, StrategyQA provides two kinds of supporting annotations for each question. The decomposed questions do not contain the answers and thus approximate the \textit{aspects} of each question. The annotated facts answers the sub-questions with accurate values, so they approximate \textit{aspects+facts}. 

In \S\ref{sec:sanity_check}, we conduct an experiment for sanity checking purposes, with results shown in Table~\ref{tab:sanity_check}. We see that T5 does not improve much when given only the \textit{aspects} (+6\%) but gains much more (31\%-37\%) when provided with \textit{aspects+facts} and additional indirect supervision. When given \textit{aspects+facts}, the model is in effect doing textual entailment. The 90.2\% accuracy shows that this entailment part of deciding how to use the facts is a much smaller bottleneck than finding the proper aspects and their values. At the same time, relatively small LMs such as T5 do not gain much from only seeing the \textit{aspects} because of their poor memorization (e.g., even if the model knows that the population of a city is needed, it cannot produce the correct number without external resources). This observation serves as the motivation for building a binary QA pipeline that first generates accurate \textit{aspects+facts} (\textbf{decompose}) and then decides the final answer with an entailment model  (\textbf{entail}).

The \textbf{decompose} step can be approached in two ways: i) generating the \textit{aspects} first, then perform information retrieval (IR) and compose a new statement for \textit{aspects+facts}; ii) generating \textit{aspects+facts} directly, then perform some factual correction because small LMs cannot memorize well. We choose the second approach for the following three reasons. 1) Our basis \modelname{} is trained on parallel news, which are natural language statements that approximate the \textit{aspects+facts} together (see Fig.~\ref{fig:overview}). 2) Generating \textit{aspects+facts} together allows the model to adhere to its beliefs and generate self-consistent logic chains because decomposition may be inter-dependent (e.g., in Fig.~\ref{fig:pipeline-overview}, the country that Cyril represents plays an important role in the next generation step). 3) Supporting facts are a much more common type of annotation (e.g., in HotpotQA) than \textit{aspects}-only annotations, which allows us to explore transfer and joint learning with other existing datasets.

\subsection{Factual Correction for Generated Facts}
\label{sec:factual-correction}

In order for generating \textit{aspects+facts} to work, we need to correct any factual errors in the generated facts. This is crucial because relatively small LMs such as T5 cannot generate accurate facts, and wrong information will hinder the performance of the entailment model when deciding the final answer. Standard information retrieval (IR) approaches aim to find a specific piece of text from a knowledge base \cite{Karpukhin2020DensePR} and tailor the correct information in the retrieved text to specific needs. However, this will not work well in our scenario because doing IR on \textit{aspects} and incorrect \textit{facts} will lead to much noise. Moreover, certain commonsense information, such as the weight of a six-year-old, are often missing from standard IR resources such as Wikipedia. 

To this end, we propose to use large-scale language models such as GPT-3 \cite{Brown2020LanguageMA} directly as a fact-checker, as we have found that GPT-3 does reasonably well on memorizing and retrieving the majority of well-known facts. Furthermore, when given appropriate prompts, GPT-3 simultaneously performs retrieval and new statement synthesis, allowing us to inspect the reasoning capability of our decomposition model directly and more efficiently. Therefore, we design a prompt that starts with \textit{``Fix the input sentence with correct facts if there are factual errors''} followed by six examples listed in Appendix~\ref{sec:appendix-prompts}. 

We emphasize that GPT-3 is only used as a fact-checker in our pipeline. It does not add any information on how to find the \textit{aspects} because it does not see the original question, rather the output of single-step generated facts. As a result, we view our ``reasoning'' component much smaller than GPT-3 as we disentangle these two parts. We discuss this more in \S\ref{sec:manual-analysis} and Appendix~\ref{sec:appendix-examples}.

\subsection{\pipelinename{} QA Pipeline}
\label{sec:pipeline}

\noindent \textbf{Decompose.} Since \modelname{} hasn't been pre-trained on questions, we fine-tune it on [question, supporting-fact] annotations from relevant datasets to generate \textit{aspects+facts} for each question. Because supporting facts are usually composed of multiple sentences, we formulate a step-by-step generation. 
For $n$ training facts, we formulate $n$ training instances from time $1$ to time $n$. At time $t$, a model sees an input sequence that is the question and all supporting facts with indices smaller than $t$ concatenated. The output sequence (learning target) is the supporting fact at index $t$. During evaluation time, the model generates one fact at a time, which then goes through the factual correction process in \S\ref{sec:factual-correction}. At time $t$, the model receives an input sequence including the original question and all current generated facts (after correction) before time $t$, and generates the $t^{th}$ supporting fact.

We design the specific input sequence as \keywordCode{[Q]Decompositions:[G(current)]}, and output sequences as \keywordCode{[G(next)]}. \keywordCode{[G(current)]} is the concatenation of all current generations, which is empty before generating the first fact. \keywordCode{[G(next)]} is the immediate next fact to be generated. 

\noindent \textbf{Entail.} With the generated facts from \textbf{decompose}, we derive binary answers for questions with the \textit{aspects+facts+indirect} model as seen in Table~\ref{tab:sanity_check}.

\subsection{Inference}
\label{sec:voting}
We sample the top five generation candidates at each generation step via diverse beam search \cite{Vijayakumar2016DiverseBS}. We select one randomly based on their $\mathrm{softmax}$ probabilities. We generate at most three facts (i.e., $t=3$ as specified in \S\ref{sec:pipeline}) or early stops for each chain if all candidates at a generation step are very similar to the current generations, determined by the SentenceBert paraphrasing model with 0.95 as the threshold. We run the three-fact generation five times for each question due to randomness in the underlying generation selection process. As a result, we will have five chains of at most three generated facts for each question. We run the entailment model individually on each chain and derive a final answer based on majority voting from each chain. The majority voting is weighted with the confidence score of the entailment model's decisions on each chains.

\section{Intrinsic Experiments}
\label{sec:intrinsic-experiments}
In this section, we conduct two intrinsic experiments with \modelname{} that directly evaluate its general decomposition capability through fine-tuning task-specific input/output sequences. We compare with T5-large as it is the base LM, and such a comparison reveals how much we improve through pre-training with parallel news distant signals. We do not compare our model with GPT-3 because few-shot learning might not be enough for it to learn the complete grammar of different tasks' decomposition. This is an advantage of fine-tuning relatively small but capable models over directly using much bigger ones in few-shot settings. All experiments use a 5e-5 learning rate, and they are repeated and averaged with three seeds.\footnote{We use $10,20,30$ as the seeds for all experiments.}

\subsection{Overnight}

\begin{table}[t]
\centering
\begin{tabular}{lccc}
\toprule
System & Hit@1 & Hit@5 & Hit@10 \\
\cmidrule(lr){1-1}\cmidrule(lr){2-2}\cmidrule(lr){3-3}\cmidrule(lr){4-4}
T5-large & 21.8 & 51.6 & 63.1 \\
\modelname{} & 48.6 & 78.9 & 85.4 \\
\bottomrule
\end{tabular}
\caption{Hit@K performances on Overnight decomposition generation. Hit@K is the percentage of instances where the top K generations contains at least one exact match. \modelname{} is from this work.}
\label{tab:overnight}
\end{table}

\noindent \textbf{Dataset and Metrics.} We evaluate and compare our model's capability to produce intermediate decomposition on the Overnight dataset \cite{Wang2015BuildingAS}. It is a semantic parsing dataset that aims to parse natural language queries into a formal parsing that can be programmatically executed to denotations. In between the natural language query and the formal parsing, it annotates an intermediate ``canonical'' form with semi-formal language, which has recently been used for work on text-based semantic parsing with transformers \cite{shin2021constrained} that we take inspiration from. For example, the annotated intermediate form of ``biggest housing unit that allows dogs'' is ``housing unit that has the largest size and that allows dogs''.
We evaluate the performance of mapping natural language queries to such intermediate forms with three domains that contain 3.8K training instances and 972 test cases. Both models are trained with three epochs. We use the same inference for both T5-large and \modelname{}, which generates ten candidates using beam search. Following previous work, the generation is also constrained by possible ``next words'', that is, we assume that we know controlled output space beforehand.

\noindent \textbf{Results and Analysis.} Table~\ref{tab:overnight} details the performance of our \modelname{} compared to its base model, T5-large. Our model doubles the performance on the exact match of the top prediction, which translates to a much higher denotation accuracy because multiple decompositions can be executed to the same denotation. Our model can find the exact match decomposition 78.9\% of the time with only five candidates to consider, showing much higher potential for end-to-end tasks that may improve through iterative learning. On the other hand, T5-large can barely cover more than half of the queries with top-five candidates and only improves to 63.1\% with more candidates (top-ten). This shows that \modelname{} is much better at making commonsense connections (e.g., ``biggest'' to ``largest size'') after fine-tuning, thanks to the pre-training process on our parallel news corpus. 

\subsection{TORQUE}

\begin{table}[t]
\centering
\begin{tabular}{lc}
\toprule
System & Exact Match \\
\cmidrule(lr){1-1}\cmidrule(lr){2-2}
T5-large & 50.3 \\
T5-large-paraphrase & 72.2 \\
\modelname{} & 82.8  \\
\bottomrule
\end{tabular}
\caption{Exact match accuracy of different models on custom-annotated TORQUE. T5-large-paraphrase is first fine-tuned on paraphrasing supervision.}
\label{tab:torque}
\end{table}

\noindent \textbf{Dataset.} TORQUE \cite{Ning2020TORQUEAR} is a temporal question-answering dataset. For example, ``what happened before...'' asks the model to find all events with a start time before that of the given event, and ``what ended before...'' should be answered with events with end times before the start time of the given event. Compared to traditional temporal relation extraction tasks, this format is more challenging to existing temporal reasoning models, as they now have to parse the question and understand what aspects (e.g., start or end times) the question is asking first. To this end, we evaluate if our proposed model can better parse the question into correct temporal phenomena.

\noindent \textbf{Annotate Decomposition.} Because TORQUE does not come with an intermediate annotation specifying the temporal properties required for each question, we need to annotate TORQUE questions with a form of intermediate decomposition to evaluate if a model understands the questions correctly. We adopt Overnight grammar for this purpose. For example, ``what started before [X]'' can be written as ``find all events whose start time is smaller than the start time of [X]''. Luckily, TORQUE uses several question templates during its annotation process. As a result, the intermediate decomposition of many questions can be automatically labeled. We create a training set of 15K question-decomposition pairs from 10 templates that are \textbf{only} about events' start time comparisons. On the other hand, we create an evaluation set of 624 questions from 11 templates, and 9 of them compare events' end times, which a model will not see during training. We do this to evaluate models' capability of ``universal'' decomposition by generalizing to unseen relations in a ``zero-shot'' fashion. For a model to do well, it must have a pre-existing representation of what the question is asking.

\noindent \textbf{Results and Analysis.} Table~\ref{tab:torque} reports the exact match accuracy on our custom TORQUE evaluation. In addition to the T5 baseline, we use the same hyper-parameters as \modelname{} to fine-tune a T5-large on the distant supervision portion from PAWS \cite{Zhang2019PAWSPA}, containing 320K sentence pairs. We do this to compare the data quality of our distant supervision and that from paraphrasing since TORQUE requires a higher level of question understanding than Overnight. All models are trained for one epoch because the training data is highly repetitive, and generate one sequence via greedy decoding. We see that our model improves more than 30\% over the baseline model, and 10\% over the paraphrasing-supervised model. More importantly, this shows that \modelname{} develops a solid implicit representation for query understanding from the pre-training, which allows it to generalize to end-time queries from supervisions that are only about start times. On the other hand, T5-large cannot correctly produce anything about end-time queries as expected.

\section{Sanity Check Experiments}
\label{sec:sanity_check}
In this section, we describe the details of the sanity check experiment mentioned and analyzed in \S\ref{sec:pipeline-intuition}.

\subsection{Dataset and Settings}

\noindent\textbf{Dataset.}
We use StrategyQA, a QA dataset with high requirements for question understanding and knowledge acquisition. It contains questions that can be answered with either ``yes'' or ``no'', and is divided into 2061/229 train/dev, and an additional 490 test questions. Each question in the training and development sets is annotated with two types of supporting evidence as shown in Fig.~\ref{fig:stqa-example}: decomposed questions and annotated facts. We use the decomposed questions as the \textit{aspects} of a question 
, and the annotated facts as \textit{aspects+facts}, as they provide specific values for the aspects. 

\noindent \textbf{Indirect Supervision.} Under the \textit{aspects+facts} setting, the model is performing general textual entailment (TE) with the given facts as the premise and the question as the hypothesis, which allows us to use indirect supervision inspired by TE. We first augment each training instance in StrategyQA with five additional instances where all supporting facts are replaced with one of their paraphrases obtained with an off-the-shelf paraphrasing model.\footnote{\url{https://huggingface.co/tuner007/pegasus_paraphrase}} We then add additional supervision from e-SNLI's development set \cite{Camburu2018eSNLINL}. We also add supervision from HotpotQA \cite{Yang2018HotpotQAAD} with its annotated supporting facts.

\subsection{Training and Results} 
We formulate a sequence-to-sequence task with input sequences as \keywordCode{[Q]Decompositions:[D]} and output sequences of either \keywordCode{yes/no}. \keywordCode{[Q]} is the question, and \keywordCode{[D]} is the additional information such as supporting facts. We fine-tune T5-3B models for three epochs under each supervision setting and evaluate with the same gold information provided during test time. Each experiment is averaged over three random seeds. Table~\ref{tab:sanity_check} details the performances on StrategyQA's development set. We have analyzed this result in \S\ref{sec:pipeline-intuition}.

\section{Decomposition QA Experiments}
\label{sec:e2e-experiments}

We detail two experiments that evaluates the QA pipeline \pipelinename{} proposed in \S\ref{sec:pipeline}. 
\subsection{Datasets} 

As argued in \S\ref{sec:pipeline-intuition}, our proposed pipeline benefits from any question-answering dataset that annotates supporting facts. To demonstrate this property, we use StrategyQA and HotpotQA jointly as supervision, and evaluate on both datasets. Because our pipeline setting is mostly designed for binary questions, we select questions that can be answered with either ``yes'' or ``no'' from HotpotQA, which accounts for 5430 questions from the training set. We use 300 binary questions from the development set of HotpotQA as evaluation. Because the supporting fact annotation in StrategyQA is human-written instead of Wikipedia sentences, it is shorter and more precise. To this end, we want the decomposition model to primarily rely on such annotations, and we duplicate each set of supporting facts in StrategyQA five times with shuffled order. These together produce around 35K decomposition instances for training.

\subsection{Settings and Baselines} 

We compare with T5-large under the same joint supervision setting (denoted as ``S+H''). We also compare with RoBERTa*-IR as described in \citet{geva2021did} on StrategyQA. It uses BoolQ \cite{clark2019boolq} as additional supervision, which is denoted as ``S+B''. We also include GPT-3 baselines, one in a regular few-shot setting and another with a few-shot chain-of-thought \cite{Wei2022ChainOT} supplement (denoted as GPT-3 CoT). Both prompts are available in Appendix~\ref{sec:appendix-prompts}. 
We report an aggregated performance (i.e., voting with all seeds as described in \S\ref{sec:voting}) on StrategyQA's development set. However, we report a single best seed's\footnote{We determine the best seed based on the StrategyQA's development set.} performance on the test set as well as HotpotQA because of both StrategyQA leaderboard's limitation and cost considerations of using GPT-3. Experiments are repeated with three random seeds, trained for three epochs with 5e-5 learning rates.

\begin{table}[t]
\centering
\small
\begin{tabular}{lcccc}
\toprule
System & Source & Dev & Test & Hotpot \\
\cmidrule(lr){1-1}\cmidrule(lr){2-2}\cmidrule(lr){3-3}\cmidrule(lr){4-4}\cmidrule(lr){5-5}
T5-Large & S+H & 55.9 & - & 56.0 \\
RoBERTa*-IR & S+B & 65.8 & 64.9 & -\\
GPT-3 & Few & 62.5 & 64.1 & 70.0 \\
GPT-3 CoT & Few & 65.9 & 63.7 & 73.0  \\
Ours & S+H & \textbf{70.3} & \textbf{67.4} & 81.0 \\
\cmidrule(lr){1-5}
Ours -pretrain & S+H & 67.2 & - & 80.7 \\
Ours -correction & S+H & 62.9 & - & 69.0 \\
Ours -joint & S or H & 65.5 & - & \textbf{81.3} \\
\bottomrule
\end{tabular}
\caption{Accuracy on StrategyQA and HotpotQA. Ours refers to the \pipelinename{} pipeline.}
\vspace{-0.2cm}
\label{tab:strategyqa_test}
\end{table}

\subsection{Results}
Table~\ref{tab:strategyqa_test} shows the performances with different baselines on StrategyQA and HotpotQA. On StrategyQA, \pipelinename{} outperforms all baseline models by 4\%, proving that our model benefits the most, and more efficiently, from existing human-annotated resources on complicated questions. On HotpotQA's binary questions, our proposed pipeline outperforms the chain-of-thought variant of GPT-3 by over 8\%, and the T5 baseline by 25\%. This shows that explicit decomposition is better than reasoning in a black box, as we achieve better performances with a decomposition model that is over 200 times smaller.\footnote{There are 770M parameters in T5-large and 175B parameters in GPT-3.}

\subsection{Ablation Studies}
We conduct ablation studies on three variants of the proposed pipeline: without the further pretraining described in \S\ref{sec:pretrain} (-pretrain), without the factual correction in \S\ref{sec:factual-correction} (-correction), and without the joint learning with both datasets (-joint). Table~\ref{tab:strategyqa_test} details the performances of ablation models. Similarly, we evaluate the ablation models on StrategyQA's development set with three random seeds and vote with all seeds, but HotpotQA only once due to cost limitations. We see that pretraining with our parallel news corpus accounts for over 3\% gain on StrategyQA. This aligns with our intuition and intrinsic experiments in \S\ref{sec:intrinsic-experiments} because StrategyQA requires advanced question understanding. Factual correction is also significant in our pipeline, which makes a 7\% difference on StrategyQA and 12\% on HotpotQA. On the other hand, joint learning contributes to the performances on StrategyQA but not on HotpotQA, which might be because HotpotQA experiments are run with single seeds.

\subsection{Manual Analysis}
\label{sec:manual-analysis}
We argue that the core of our improvement is producing proper decompositions instead of the use of GPT-3. We conduct a manual analysis on 20 questions\footnote{We use the first 20 questions in the dev set that have agreeable annotated facts, without looking at the predictions.} from StrategyQA's dev set and inspect the raw decomposition before factual correction. We find that \modelname{} fails to produce at least one decomposition with all necessary aspects on only two. This suggests that \modelname{} does well in understanding ~90\% of the questions without GPT-3, even though we need factual correction for the entailment model to produce the correct answer. Moreover, the analysis shows that GPT-3 does not provide anything beyond correcting any factual errors in the statement generated by \modelname{}, as it only sees one decomposition at a time without seeing the actual question. We provide some actual output examples in Fig.~\ref{fig:manual-analysis-examples} for more insights.

\section{Conclusion}
This work proposes a novel method that extracts distant and incidental signals from parallel news to facilitate general question representation. Such parallel news signals intuitively bridge the reasoning gap in pre-trained language models due to reporting biases. To support this intuition, we train a model named \modelname{} on such distant supervision and show that it improves 20\%-30\% on two semantic parsing benchmarks, namely Overnight and TORQUE, that directly evaluate query understanding. With \modelname{} as the basis, we design a well-motivated question-answering pipeline \pipelinename{} that follows a decomposition, correction, and entailment scheme. We show that \pipelinename{} improves on StrategyQA and HotpotQA by 3.7\% and 8\%, respectively.

\section*{Acknowledgments}
This research is based upon work supported in part by the office of the Director of National Intelligence (ODNI), Intelligence Advanced Research Projects Activity (IARPA), via IARPA Contract No. 2019-19051600006 under the BETTER Program, and by Contract FA8750-19-2-1004 with the US Defense Advanced Research Projects Agency (DARPA). The views expressed are those of the authors and do not reflect the official policy or position of the Department of Defense or the U.S. Government. We also thank the Aristo team at the Allen Institute for AI for valuable support and feedback throughout the entire research process.

\section{Limitations}
In this section, we discuss some of the limitations of our work, and motivate future works.

\noindent \textbf{Limited Question Formats.} Our proposed QA pipeline operates on binary \textit{yes/no} questions. While binary questions are very general, as most other questions can be re-written into similar forms, such transformations have not been designed or evaluated, which motivates future works.

\noindent \textbf{Limited Factual Correction Coverage.} We use GPT-3 as the backbone for our factual correction step. Although it is shown to be effective, it is not as deterministic as Wikipedia-based IR approaches, and we cannot easily interpret why it makes mistakes and understand how to improve.

\bibliography{anthology,custom}
\bibliographystyle{acl_natbib}

\clearpage

\appendix

\section{GPT-3 Prompts}
\label{sec:appendix-prompts}
The prompts for factual correction is shown in Table~\ref{tab:prompt_correction}.
For QA, we use same prompts in \cite{Wei2022ChainOT}. We list the prompts with binary answers in Table \ref{tab:prompt_binary}, and the prompts with chain of thought in Table \ref{tab:prompt_cot}.

\begin{table}[H]
\centering
\small
\begin{tabular}{l}
\toprule
Fix the input sentence with correct facts if \\ there are factual errors. \\ \\
\textbf{Wrong:} Mount Fuji is in China. \\
\textbf{Correct:} Mount Fuji is in Japan.\\\\
\textbf{Wrong:} Amy Winehouse was diagnosed with \\stage 4 breast cancer in May 2017. \\
\textbf{Correct:} Amy Winehouse was not diagnosed \\ with cancer. \\\\
\textbf{Wrong:} Ten gallons of seawater weigh 650 pounds. \\
\textbf{Correct:} Ten gallons of seawater weigh \\ approximately 83 pounds. \\\\
\textbf{Wrong:} Buffalo wings contain capsaicin. \\
\textbf{Correct:} uffalo wings contain capsaicin. \\\\
\textbf{Wrong:} The Albany in Georgia has over 50,000 people. \\
\textbf{Correct:} The Albany in Georgia has over 73,000 people. \\\\
\bottomrule
\end{tabular}
\caption{Factual correction prompt.}
\label{tab:prompt_correction}
\end{table}

\begin{table}[H]
\centering
\small
\begin{tabular}{l}
\toprule
\textbf{Q:} Do hamsters provide food for any animals?\\
\textbf{A:} Yes.\\\\
\textbf{Q:} Could Brooke Shields succeed at University of \\
Pennsylvania?\\
\textbf{A:} Yes.\\\\
\textbf{Q:} Yes or no: Hydrogen's atomic number squared \\
exceeds number of Spice Girls?\\
\textbf{A:} No.\\\\
\textbf{Q:} Yes or no: Is it common to see frost during \\
some college commencements?\\
\textbf{A:} Yes.\\\\
\textbf{Q:} Yes or no: Could a llama birth twice during \\
War in Vietnam (1945-46)?\\
\textbf{A:} No. \\\\
\textbf{Q:} Yes or no: Would a pear sink in water?\\
\textbf{A:} No.\\
\bottomrule
\end{tabular}
\caption{Few-shot prompt with binary answers only.}
\label{tab:prompt_binary}
\end{table}

\begin{table}[H]
\centering
\small
\begin{tabular}{l}
\toprule
\textbf{Q:} Do hamsters provide food for any animals?\\
\textbf{A:} Hamsters are prey animals. Prey are food \\
for predators. Thus, hamsters provide food for\\
some animals. So the answer is yes.\\\\
\textbf{Q:} Could Brooke Shields succeed at University of \\
Pennsylvania?\\
\textbf{A:} Brooke Shields went to Princeton University.\\
Princeton University is about as academically\\
rigorous as the University of Pennsylvania. \\
Thus, Brooke Shields could also succeed at the \\
University of Pennsylvania. So the answer is yes.\\\\
\textbf{Q:} Yes or no: Hydrogen's atomic number squared \\
exceeds number of Spice Girls?\\
\textbf{A:} Hydrogen has an atomic number of 1. 1 squared\\
is 1. There are 5 Spice Girls. Thus, Hydrogen’s atomic \\
number squared is less than 5. So the answer is no.\\\\
\textbf{Q:} Yes or no: Is it common to see frost during \\
some college commencements?\\
\textbf{A:} College commencement ceremonies can happen in \\
December, May, and June. December is in the winter, \\
so there can be frost. Thus, there could be frost at \\
some commencements. So the answer is yes.\\\\
\textbf{Q:} Yes or no: Could a llama birth twice during \\
War in Vietnam (1945-46)?\\
\textbf{A:} The War in Vietnam was 6 months. The gestation\\
period for a llama is 11 months, which is \\
more than 6 months. Thus, a llama could not \\
give birth twice during the War in Vietnam. \\
So the answer is no.\\\\
\textbf{Q:} Yes or no: Would a pear sink in water?\\
\textbf{A:} The density of a pear is about 0.6g/cm3,\\
which is less than water. Objects less dense \\
than water float. Thus, a pear would float. \\
So the answer is no.\\
\bottomrule
\end{tabular}
\caption{Few-shot prompt with chain of thought.}
\label{tab:prompt_cot}
\end{table}

\section{Decomposition Examples}
\label{sec:appendix-examples}
In this section, we list 5 questions from StrategyQA and corresponding decompositions from \pipelinename{}.

\begin{figure*}
    \centering
    {\footnotesize
            \begin{tabular}{| p{2cm} | p{3.2cm} | p{3.2cm} | p{3.2cm} |}
                 \hline 
                 \multicolumn{1}{|c}{\textbf{Question}} & \multicolumn{1}{c}{\textbf{Gold Decomposition}} & \multicolumn{1}{c}{\textbf{\pipelinename{} Decomposition}} & \multicolumn{1}{c|}{\textbf{\pipelinename{}  w/o correction}}\\ \hline
                 \emph{Will the Albany in Georgia reach a hundred thousand occupants before the one in New York?} & Albany, GA has around 75,000 people ; Albany, NY has almost 100,000 people & The Albany in Georgia has over 73,000 people. ; The Albany in New York has over 97,000 people. ; The Albany in Georgia is not the largest city in Georgia. &  The Albany in Georgia has a population of \hlerror{63,417}. ; The Albany in New York has a population of 97,050. ; There are over a hundred thousand people living in New York. \\ \hline
                 \emph{Could Lil Wayne legally operate a vehicle on his own at the beginning of his career?} & Lil Wayne's career began in 1995, at the age of 12, when he was signed by Birdman and joined Cash Money Records as the youngest member of the label. ; A driver's license is required to legally operate a vehicle by yourself in the USA. ; The eligible age to first obtain a driver's license varies substantially from state to state, from 14 years, three months, in South Dakota to 17 in New Jersey. & Lil Wayne was born in 1982. ; Lil Wayne started his career in 1996. ; The minimum age to operate a vehicle is 16 years old. & Lil Wayne was born in \hlerror{1991}. ; Lil Wayne began his career in \hlerror{1996}. ; The minimum age to operate a vehicle is \hlerror{21 years old}. \\ \hline
                 \emph{Are you likely to find a crucifix in Karachi?} & The crucifix is a symbol of Christianity ; The vast majority of Pakistan's population is Muslim & The Karachi area of Pakistan is a predominantly Muslim city. ; A crucifix is a statue of Jesus that is placed on a cross. ; The majority of the population of Karachi is Muslim. & The Karachi area of Pakistan is a predominantly Muslim city.. ; A crucifix is a statue of Jesus that is placed in a church.. ; Jesus is the \hlerror{patron saint} of Muslims.\\ \hline
                 \emph{Would George Fox support stoning?} & George Fox was the founder of the Religious Society of Friends, commonly known as the Quakers or Friends. ; The Quakers advocate for peace and nonviolence. ; Stoning is a particularly violent and brutal method of capital punishment. &  George Fox was a famous Quaker character. ; Stones are not used to punish people for their sins. ; Quakers are a religious group that opposes the use of violence. & George Fox was a famous \hlerror{Fox character}. ; Stones are used to punish offenders for their crimes. ; George Fox was an American politician. \\ \hline
                 \emph{Can Cyril Ramaphosa become Secretary General of NATO?} & Cyril Ramaphosa is the President of South Africa ; The Secretary General of NATO comes from one of NATO's member countries ; South Africa is not a member of NATO & Cyril Ramaphosa is the President of South Africa. ; The Secretary General of NATO is Jens Stoltenberg. ; South Africa is not a member of NATO. & Cyril Ramaphosa is the current President of South Africa. ; South Africa \hlerror{is a member of} NATO. ; The Secretary General of NATO is \hlerror{John Bolton} \\ \hline
            \end{tabular}}
    \caption{Examples of StrategyQA questions, gold decompositions, \pipelinename{} decompositions and \pipelinename{} decompositions without factual correction from GPT-3. \hlerror{Errors} in the decomposition without correction are highlighted. Due to randomness, the decompositions with and without factual correction do not form 1-to-1 mapping. We see that \pipelinename{} decompositions are very good on many occasions, and even without the help of GPT-3 and factual correction, the decompositions demonstrate good question understanding.} 
    \label{fig:manual-analysis-examples}
\end{figure*}

\end{document}